\documentclass{article}
\usepackage{ijcai24}
\usepackage{times}
\usepackage{soul}
\usepackage{url}
\usepackage[hidelinks]{hyperref}
\usepackage[utf8]{inputenc}
\usepackage[small]{caption}
\usepackage{graphicx}
\usepackage{amsmath}
\usepackage{amsthm}
\usepackage{booktabs}
\usepackage{algorithm}
\usepackage{algorithmic}
\usepackage[switch]{lineno}
\usepackage{marvosym}  
\usepackage{amssymb}  
\nolinenumbers

\title{Semantic Localization Guiding Segment Anything Model For Reference Remote
Sensing Image Segmentation}

\author{
Shuyang Li$^1$
\and
Shuang Wang$^{1}$\and
Tao Xie$^{1,2}$\And
Zhuangzhuang Sun$^1$\\
\affiliations
$^1$School of Artificial Intelligence, Xidian University,Xi’an, Shaanxi Province, China,\\
$^2$Shaanxi Satellite Application Center for Natural Resources, Shaanxi, China\\
\emails
}
\begin{document}

\maketitle
\begin{abstract}
  The Reference Remote Sensing Image Segmentation (RRSIS) task generates segmentation masks for specified objects in images based on textual descriptions, 
which has attracted widespread attention and research interest. 
Current RRSIS methods rely on multi-modal fusion backbones and semantic segmentation heads 
but face challenges like dense annotation requirements and complex scene interpretation.
To address these issues, 
we propose a framework named \textit{prompt-generated semantic localization guiding Segment Anything Model}(PSLG-SAM), 
which decomposes the RRSIS task into two stages: 
coarse localization and fine segmentation. 
In coarse localization stage, a visual grounding network roughly locates the text-described object. 
In fine segmentation stage, 
the coordinates from the first stage guide the Segment Anything Model (SAM), enhanced by a clustering-based foreground point generator and a mask boundary iterative optimization strategy for precise segmentation. 
Notably, the second stage can be train-free, significantly reducing the annotation data burden for the RRSIS task. 
Additionally, 
decomposing the RRSIS task into two stages allows for focusing on specific region segmentation, avoiding interference from complex scenes.We further contribute a high-quality, multi-category manually annotated dataset.
Experimental validation on two datasets (RRSIS-D and RRSIS-M) demonstrates that PSLG-SAM achieves significant performance improvements and surpasses existing state-of-the-art models.Our code will be made publicly available.
\end{abstract}




\maketitle

\section{Introduction}
\begin{figure}[!htpb]
    \centering
    \includegraphics[width=\linewidth]{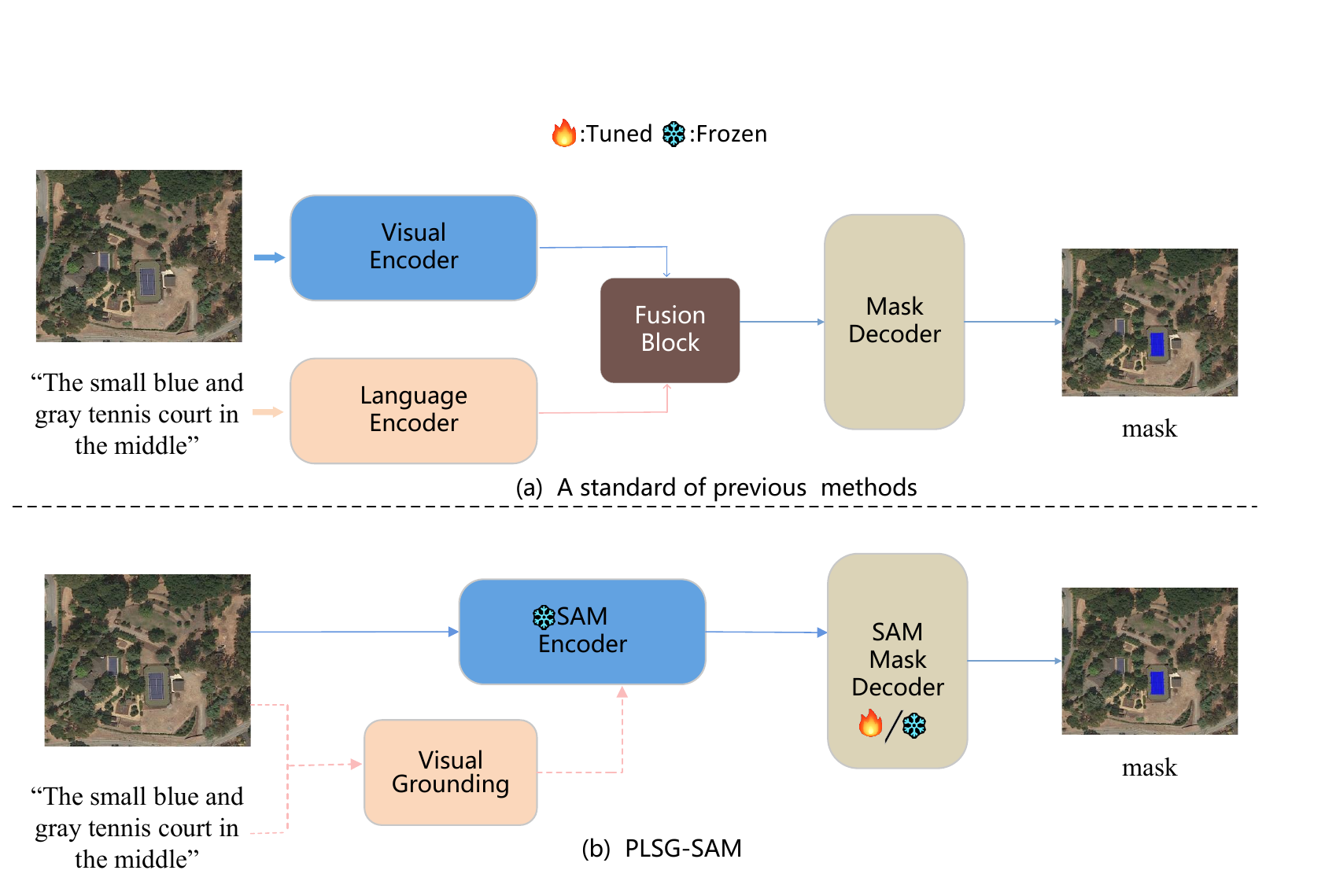}
    \caption{In comparison to the proposed PSLG-SAM with traditional RRISS methods, we introduce a visual grounding model that transforms referring segmentation into a two-step process. In the first step, the visual grounding model converts semantic cues into coordinate prompts. In the second step,we use SAM to segment the target object.}
    \label{fig:overall}
\end{figure}
Reference Remote Sensing Image Segmentation (RRSIS) is an emerging task in the field of remote sensing image processing.
It integrates reference expressions with remote sensing images to predict the mask of target entities. 
Specifically, the model generates pixel-level segmentation masks for objects in a given image based on textual descriptions.
This task holds significant potential to drive advancements across various fields,
such as urban land planning \cite{pham2011case,wellmann2020remote}, fisheries monitoring \cite{klemas2013fisheries}, and climate change detection \cite{yang2013role}. Currently, methods in RRSIS are mainly focused on improving segmentation networks based on dual encoders. 
The  \cite{yuan2024rrsis}first introduced the RRSIS task and proposed a dataset specific to this field. 
They analyzed the challenges of applying natural image-based reference image segmentation methods to remote sensing field.
A semantic interaction scale-enhanced network was designed to improve the segmentation performance of scattered targets in remote sensing images in this work.
Another study \cite{liu2024rotated} created a higher-quality benchmark dataset and proposed a rotated multi-scale interaction network for the segmentation of rotated targets. 
These single-stage methods have made notable progress in the field.

However, compared to natural image RIS tasks, 
RRSIS deals with smaller target objects and typically more complex background regions. 
Additionally, the limited and challenging annotation of region description datasets on remote sensing images makes it difficult to train models to achieve the required accuracy for this task.

To address these issues, we propose a new framework named PSLG-SAM, as shown in \autoref{fig:overall}. This framework decomposes the RRSIS task into two stages:
coarse localization and fine segmentation. 
In the coarse localization stage, we leverage the rich spatial relationships and object attributes described in the text to determine the target's location within the image.
In the fine segmentation stage, 
the framework leverages SAM's zero-shot boundary segmentation capability and the target's location information 
to achieve precise segmentation.
Our two-stage framework utilizes SAM's boundary segmentation capability for fine-grained segmentation, 
effectively circumventing the challenge of directly generating precise masks from textual descriptions. 
Moreover, The second stage can be performed without training, greatly alleviating the annotation data burden for the RRSIS task.

Nonetheless, the bounding boxes generated by visual grounding models are often imprecise, either including only partial targets or being excessively large relative to the objects. 
To address this, we design an unsupervised cluster centroid generator that captures the geometric distinctions between the foreground and background, producing center foreground point prompts for SAM. 
Given the complexity of object interiors, 
regions with diverse attributes (e.g., color, texture, or shape) may exist, 
potentially hindering the model's ability to accurately delineate object boundaries. 
Therefore, we employ a simple yet effective iterative strategy inspired by the GrabCut algorithm \cite{rother2004grabcut} to refine and optimize the mask boundaries. 
This ensures the model comprehensively captures the object's edge features. 

To validate the effectiveness of the proposed method, we have constructed a referring image segmentation dataset specifically designed for remote sensing imagery. This dataset exhibits multi-scale and multi-category characteristics while maintaining superior mask quality. A comprehensive description of this dataset will be presented in \autoref{4.1}.
\par The main contributions can be summarized as follows:
\begin{itemize}
\item We propose a framework that decomposes the RRSIS task into sequential visual grounding and region segmentation tasks. To enable comprehensive evaluation, we introduce a new multi-scale benchmark dataset with expert-annotated masks. The framework performs excellently under both weakly supervised and standard experimental settings on both the standard RRSIS-D benchmark and our newly proposed dataset.
\item We design an unsupervised centroid generator that adds precise center-point prompts to rough bounding box hints. The generated foreground centroids effectively aassist the decoder in focus on the target object,thereby enhancing segmentation accuracy.
\item Additionally, a mask boundary iterative optimization strategy is employed to refine and enhance the prediction results of foreground edges. This strategy effectively helps capture the features of boundary regions, thereby improving the overall prediction performance.
\end{itemize}
\section{Related Work}
\subsection{Referring Remote Sensing Image Segmentation}
\par Despite significant advancements in the task of referring image segmentation 
in natural images domain~\cite{li2018referring,ye2019cross,liu2017recurrent,margffoy2018dynamic,yu2018mattnet}, 
The task of referring image segmentation has made significant progress in the field of processing natural images, but its application to remote sensing image remains an emerging area still in its early stages. Yuan et al. \cite{yuan2024rrsis} introduced the first remote sensing indicative image segmentation dataset, which contains 4,420 image-language-label triplets. They also proposed a multi-scale interactive feature extraction module based on LAVT \cite{yang2022lavt}, which enhanced the segmentation ability for small objects. In \cite{liu2024rotated}, a novel rotated multi-scale interaction network (RMSIN) was presented, along with a more complex and diverse dataset, RRSIS-D, to address existing gaps in referring remote sensing image segmentation. RMSIN was designed on top of the LAVT framework, incorporating an intra-scale interaction module and a cross-scale interaction module to promote comprehensive feature fusion. Additionally, RMSIN’s decoder integrates adaptive rotation convolutions, which significantly improve the model's performance in segmenting rotated objects. This integration enhances the network's ability to capture and represent the rotation of complex objects, thereby improving the overall performance.However, both of two existing methods are limited by the scarcity of labeled remote sensing images. Therefore, we propose combining SAM and a visual grounding model to replace the end-to-end segmentation approach.
\subsection{Visual Grounding on RS Images}
Compared to Visual Grounding algorithms based on natural images\cite{hu2016natural,yu2018mattnet,ye2022shifting}, the field of Visual Grounding for remote sensing (RS) images has started relatively late. The scenes in RS images are more complex, requiring visual models to focus more on the geographical relationships between objects. Sun et al \cite{sun2022visual} were the first to introduce the Remote Sensing Visual Grounding (RSVG) task and collected a dataset called RSVG for this task. They designed the GeoVG model to capture local relationships within RS images. Zhan et al \cite{zhan2023rsvg}also proposed a new dataset, DIOR-RSVG, and introduced the MGVLF framework for the RSVG task. This framework, based on the TransVG model \cite{deng2021transvg}, integrates multi-scale visual features and multi-granularity text features through a multi-granularity cross-modal module. Inspired by the Detection Transformer (DETR) \cite{carion2020end}architecture, another approach, LQVG, was proposed, which utilizes a multi-scale cross-modal alignment (MSCMA) module to enhance the semantic relevance between cross-modal features. Due to its advanced grounding performance, our framework adopts LQVG as the model for generating coordinate prompts.
\section{Method}
\begin{figure*}[!ht]
    \centering
    \includegraphics[width=1\linewidth]{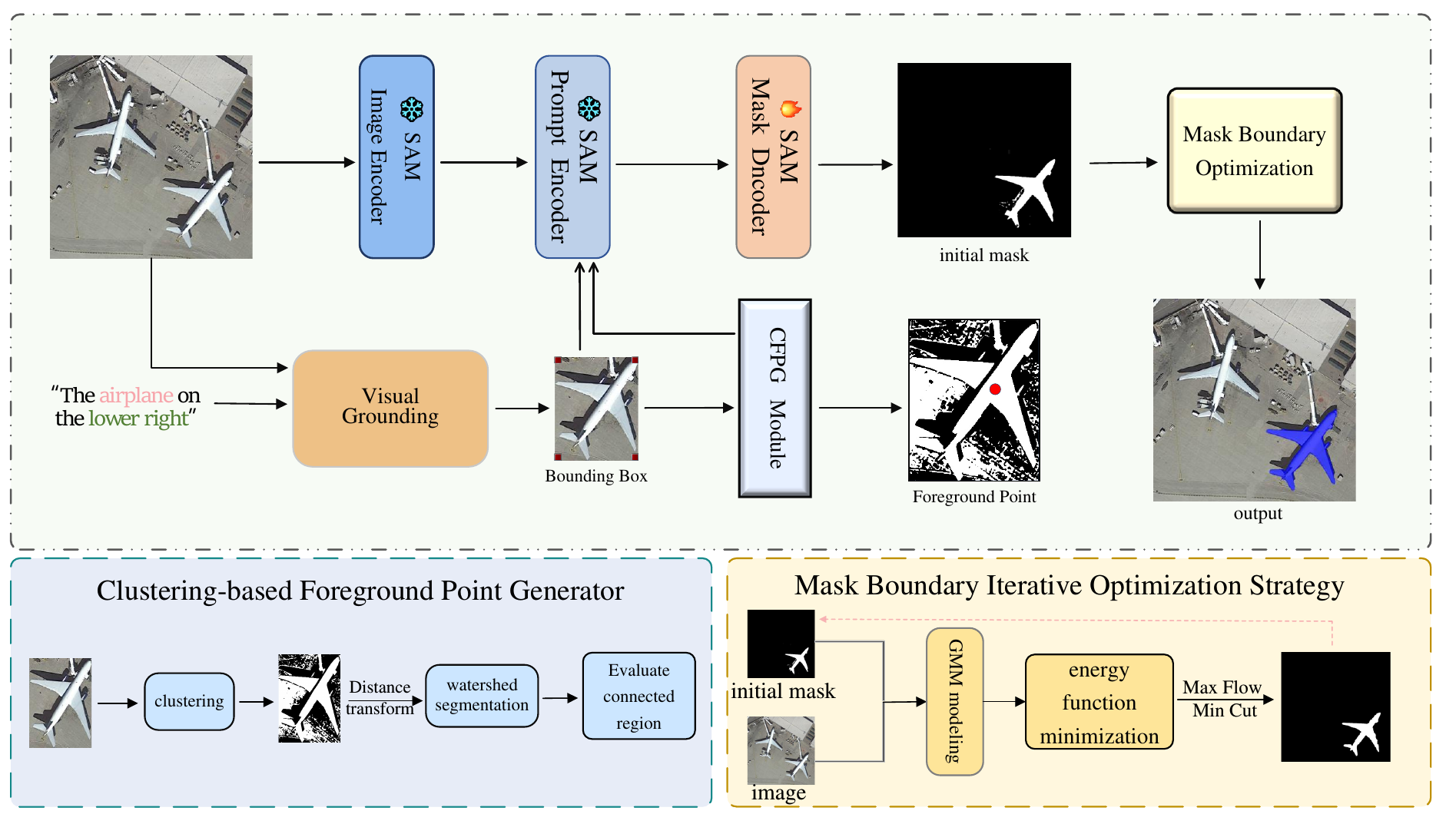}
    \caption{The pipeline of proposed PSLG-SAM. A visual grounding model is introduced to transform the reference image segmentation into localization and region segmentation. The proposed Clustering-based Foreground Point Generator provides additional foreground prompts to SAM. The mask boundary iterative optimization strategy (Mask Boundary
Optimization) enhances edge segmentation precision through iterative refinement.}
    \label{pipeline}
\end{figure*}
\subsection{Overview}
The proposed PSLG-SAM framework operates as illustrated in \autoref{pipeline}. It takes remote sensing image $I \in R^{H \times W \times 3}$ and textual expression $T \in R^{1 \times d}$ as input. The framework comprises three main components: a vision grounding model tailored for remote sensing images to generate spatial features, the SAM visual encoder (including a frozen image encoder and a prompt encoder), and a lightweight decoder responsible for producing segmentation masks. Additionally, the framework incorporates two auxiliary modules—namely, the clustering-based foreground point generator (CFPG) and a mask boundary iterative optimization strategy(MBO)—designed to further enhance segmentation accuracy.It is worth noting that when performing segmentation with the mask decoder at the end, we have the option to either fine-tune the decoder or keep it frozen. In this regard, our framework can be viewed as a weakly supervised approach, as we do not use mask-level annotated data.

\noindent \textbf{\textit{\resizebox{0.175\textwidth}{!}{Visual grounding.}}}
 The generic vision grounding model comprises two encoders and a decoder. Given a three-channel input image $I$ and textual input $T$, the two encoders separately extract visual features $F_v $ and textual features $F_t $. Notably, the extracted visual features are coarse and cannot be directly used for fine-grained segmentation. Subsequently, $F_v$ and $F_t$ are fused to produce bounding box coordinates as follows:
\begin{equation}
F_C = \text{Dec}_{\text{vg}}(F_v \odot \phi(F_t))
\label{eq:fusion}
\end{equation}
where $\phi$ represents a mapping operation, and $\odot$ denotes the fusion operation. The resulting coordinates $F_C$ are then used as part of the input to the prompt encoder. The selection of the Visual Grounding Model is quite flexible, and we adopt LQVG to generate coordinate representations.

\noindent \textbf{\textit{\resizebox{0.21\textwidth}{!}{Frozen visual encoder.}}} The SAM visual encoder adopts a Vision Transformer (ViT) architecture, incorporating both local window attention and periodic global attention. This encoder demonstrates strong generalization capability to unseen images. Given the three-channel input image $I \in R^{H \times W \times 3}$, the frozen visual encoder generates global image features $E_I$ with consistent feature dimensions. To reduce computational and training overhead, we utilize only the final-layer visual features instead of multi-scale features. Meanwhile, the prompt encoder takes the bounding box coordinates $F_c$ from the vision grounding model and point prompts from the auxiliary modules as input, generating prompt features:
\begin{equation}
E_p = \text{Enc}_p(F_C, D(x, y))
\label{eq:prompt_features}
\end{equation}
where $D(x, y)$ represents the point coordinate produced from CFPG, which denotes the center of the foreground target object.

\noindent \textbf{\textit{\resizebox{0.145\textwidth}{!}{Mask dencoder.}}}After obtaining $E_i$ and $E_p$, they are fed into the lightweight decoder to produce the final segmentation output:
\begin{equation}
F_o = \text{Dec}_{\text{SAM}}(E_I , E_p)
\label{eq:final_output}
\end{equation}
Here, the fusion operation in lighting decoder is implemented through multiple cross-attention layers.After the decoder outputs $F_o$, we  apply the MBO strategy to improve the edge prediction. Since SAM employs a lightweight decoder, we explored fine-tuning the decoder and observed that using limited data and computational resources  significantly improves the final results. 

\subsection{Clustering-based Foreground Point Generator}

The clustering-based foreground point generator is designed to produce accurate foreground point hints for object detection or segmentation tasks. The method first crops the image based on coordinates, then applies clustering segmentation to generate a binary image, and finally performs connected component analysis on the binary image to extract the target points. The detailed workflow is described below.

Given an input image $I \in {R}^{H \times W \times 3}$ and bounding box coordinates $[x_1, y_1, x_2, y_2]$, the region of interest (ROI) is first extracted. The cropping operation is defined as:
\begin{equation}
I_{\text{ROI}} = I[y_1:y_2, x_1:x_2]
\label{eq:roi_crop}
\end{equation}
The cropped image contains only pixels relevant to the target region.Within the extracted ROI, we use clustering operations to generate the foreground region.
The preprocessed image is then flattened into a 2D feature matrix $P \in {R}^{N \times 3}$, where $N = (y_2 - y_1) \times (x_2 - x_1)$ is the total number of pixels.
Next, KMeans++ \cite{2007K} is applied to the pixel features to divide them into $k=2$ classes (foreground and background). Such simple clustering may group the foreground with part of the background into one class; we will filter out the foreground in the subsequent steps. The clustering objective is to minimize the sum of squared distances within clusters:
\begin{equation}
\arg \min \sum_{i=1}^N \sum_{j=1}^k w_{ij} \|P_i - C_j\|^2
\label{eq:kmeans_objective}
\end{equation}
where $C_j$ denotes the centroid of the $j$-th cluster, and $w_{ij}$ is an indicator function that equals 1 if pixel $P_{i}$ belongs to cluster $C_j$, and 0 otherwise.The clustering results are converted into a binary segmentation map $I_{\text{ROI}}^{'}$.

Subsequently, morphological operations are applied to $I_{\text{ROI}}^{'}$ to eliminate noise and bridge small gaps by performing opening and closing operations. At this stage, certain background regions in the binary map may be assigned the same label as the target foreground region due to the complex geographical relationships in remote sensing images.Therefore, it is necessary to distinguish the background from the target foreground region.

To address this, watershed segmentation\cite {785528} is applied to binary segmentation map.The process begins with a distance transform that calculates the shortest distance from each foreground pixel to the nearest background pixel:
\begin{equation}
D(x, y) = \min_{(x', y') \in \text{Background}} \| (x, y) - (x', y') \|.
\end{equation}
Foreground markers are generated by thresholding the distance transform with a threshold,which ensures that only relevant foreground regions are marked.The watershed algorithm is then applied using these markers, segmenting the image into distinct regions, with boundaries labeled as $-1$.Consequently, we obtain $I_{\text{ROI}}^{''}$, which comprises several connected regions with distinct separation between the background and foreground.
 
Although the foreground and background have been separated, it remains unclear which specific region or regions correspond to the target object. Upon conducting geometric analysis of the image, we observed that almost all foreground objects have smaller convexity defects than the background regions.Therefore, we only need to find the connected region with the greatest convexity,which can be regarded as part of the foreground object..In the $I_{\text{ROI}}^{''}$, connected component analysis is performed to extract all continuous foreground regions. Let $R_i$ denote the $i$-th connected region, 
and the set of all connected regions is expressed as:$\{R_1, R_2, \dots, R_n\}$
Each connected component is evaluated for its area $A_(R_i)$ and convexity $\kappa_i$, where convexity is defined as the ratio of the region's area to its convex hull area:
\begin{equation}
\text{$\kappa_i$} = \frac{\text{A}(R_i)}{A_{\text{hull}, (R_i)}}
\label{eq:region_convexity}
\end{equation}
Regions are filtered based on a threshold for area, and the region with the highest convexity is selected as the optimal region:
\begin{equation}
R_{\text{best}} = \arg \max_{R_i} \{\text{$\kappa_i$} \,|\, \text{Area}(R_i) > \text{Threshold}\}
\label{eq:best_region}
\end{equation}
For the selected optimal region $R_{\text{best}}$, its geometric center is calculated as the final foreground point:
\begin{equation}
(x_c, y_c) = \frac{1}{|R_{\text{best}}|} \sum_{(x, y) \in R_{\text{best}}} (x, y)
\label{eq:foreground_point}
\end{equation}
where $|R_{\text{best}}|$ represents the total number of pixels in $R_{\text{best}}$. The computed foreground point $(x_c, y_c)$ is the foreground center point we obtained.

\subsection{Mask Boundary Iterative Optimization Strategy}
We implemented an effective strategy to enhance edge segmentation performance, 
inspired by the GrabCut algorithm. Given the initial mask \( F_o \), 
an erosion operation is applied to the mask to eliminate noise. The eroded mask is then treated as the foreground region, 
with the remaining part considered as the background region.
Next, we construct Gaussian Mixture Models (GMMs) for both foreground and background, 
with their parameters initialized using maximum likelihood estimation. For each pixel \( Z_i \), 
the probability of it belonging to each Gaussian component of the foreground and background is computed using the Gaussian Mixture Model. 
Specifically, the foreground probabilities are calculated as:
\begin{equation}
P_F(Z_i) = \sum_{l=1}^{K} w_{l,fg} \cdot \mathcal{N}(Z_i \mid \mu_{l,fg}, \Sigma_{l,fg})
\end{equation}
where \( P_F(Z_i) \) represent the probabilities that pixel \( Z_i \) belongs to the foreground, respectively, and \( \mathcal{N}(Z_i \mid \mu_{l,fg}, \Sigma_{l,fg}) \) are the probability density functions of the \( l \)-th Gaussian component for the foreground , respectively. The probability calculation for the background $P_B(Z_i)$ is the same as for the foreground. These probabilities are then used to calculate the posterior probabilities \( \gamma_{i,j}^{fg} \) and \( \gamma_{i,j}^{bg} \) that pixel \( Z_i \) belongs to each Gaussian component in the foreground and background, using Bayes' inference formula:
\begin{equation}
\gamma_{i,j}^{fg} = \frac{w_j^{fg} \mathcal{N}(Z_i \mid \mu_j^{fg}, \Sigma_j^{fg})}{P_F(Z_i)}
\end{equation}
and
\begin{equation}
\gamma_{i,j}^{bg} = \frac{w_j^{bg} \mathcal{N}(Z_i \mid \mu_j^{bg}, \Sigma_j^{bg})}{P_B(Z_i)}
\end{equation}
These posterior probabilities are normalized by the total probability of the foreground and background, and are used to evaluate the connection strength between the pixels and foreground or background. Then the optimization is formulated as an energy minimization problem, where the objective function consists of two terms: the data term, which measures the connection strength between pixels and foreground or background, and the smoothness term, which measures the similarity between adjacent pixels. The data term is calculated by taking the negative logarithm of the probability of each pixel belonging to the foreground or background and summing them up.The smoothness term is given by the similarity between neighboring pixels, often based on color difference, texture similarity, or spatial proximity. 

The final objective function is the weighted sum of these two terms. To minimize this objective function, we use the Max-Flow Min-Cut algorithm, which optimizes the pixel labels and achieves the optimal partitioning of the foreground and background. After each optimization step, the Gaussian Mixture Model parameters for the foreground and background are updated using the Expectation-Maximization (EM) algorithm: in the E-step, the posterior probabilities \( \gamma_{i,j}^{fg} \) and \( \gamma_{i,j}^{bg} \) are computed based on the current labeling, and in the M-step, the model parameters (mean, covariance, and weight) are updated according to these probabilities. This iterative process continues until the segmentation converges.

\section{Experiment}
\subsection{RRSIS-M Dataset}
\label{4.1}
As an emerging domain, RRSIS continues to face challenges due to the scarcity of publicly available datasets. Existing benchmarks, such as RefSeg, predominantly feature word-level referring expressions but lack fine-grained semantic constraints for delineating target objects. Similarly, the RRSIS-D dataset, which leverages the Segment Anything Model for semi-automatic segmentation, still exhibits significant limitations, primarily manifested as coarse mask boundaries and inconsistent category labeling.

\begin{figure*}[!h]
    \centering
    
    \includegraphics[width=1 \linewidth]{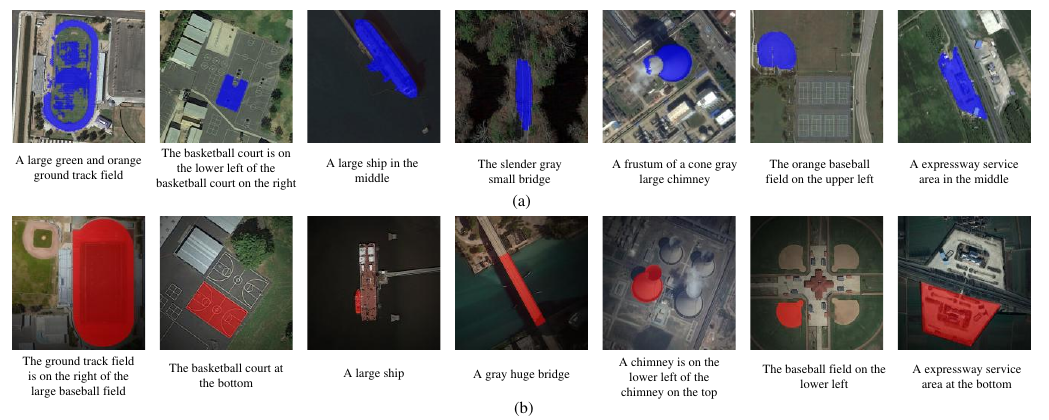}
    \caption{Qualitative comparison of mask annotation quality between RRSIS-D (a) and RRSIS-M (b).
    }
    \label{dataset}
\end{figure*}

To overcome these shortcomings, we introduce \textbf{RRSIS-M}, a high-quality benchmark meticulously crafted for RRSIS tasks. In contrast to RefSeg and RRSIS-D, RRSIS-M offers superior mask precision through a rigorous, professionally supervised annotation process. The construction of RRSIS-M follows a structured pipeline, detailed as follows:
    \begin{enumerate}
        \item[] \hspace*{-2em}  {Step 1. Initial Annotation}: Leveraging bounding box prompts and textual descriptions derived from the DIOR-RSVG dataset, initial segmentation masks are generated via expert manual annotation.
        \item[] \hspace*{-2em} {Step 2. Quality Assurance}: To ensure annotation consistency and reliability, a bipartite quality control process is implemented:
        \begin{itemize}
            \item \textit{Domain-Knowledge Filtering}: Images with ambiguous or inadequately described targets are excluded based on expert domain knowledge.
            \item \textit{Class Balancing}: Supplementary annotations are incorporated to address underrepresented categories, ensuring a balanced class distribution.
        \end{itemize}
        \item[] \hspace*{-2em} {Step 3. Dataset Partitioning}: The dataset is divided into a training set (4,361 images) and a test set (1,104 images), with the test set designed to maintain an equitable distribution of instances across categories. A secondary review eliminates any residual labeling errors.
    \end{enumerate}
The resulting RRSIS-M dataset comprises 5,465 high-resolution remote sensing images ($800\times800$ pixels) spanning 20 semantic categories.Unlike RRSIS-D, which adopts a selective annotation strategy retaining only a single instance per scene, RRSIS-M preserves multiple object instances within individual scenes, thus maintaining real-world complexity. The mask annotations in RRSIS-M demonstrate significantly superior overall accuracy and edge segmentation quality compared to existing benchmarks, particularly in delineating boundaries between referred objects and background. Visual comparisons with RRSIS-D are provided in Figure~\ref{dataset}. To facilitate integration with prevailing referring segmentation frameworks, annotations are provided in both RefCOCO format and as mask-image pairs.
\subsection{Implementation Details}
Our framework consists of two components: visual grounding model and Segment Anything Model.For visual grounding model, we use a ResNet50\cite{he2016deep} pre-trained on ImageNet as the image encoder and BERT\cite{devlin2018bert} as the text encoder. The model is optimized using the AdamW optimizer, with an initial learning rate of $1 \times 10^{-5}$ for the text encoders and $1 \times 10^{-4}$ for the other components. We trained the model for 60 epochs, with a learning rate decay factor of 0.1 applied at epoch 45. The batch size was set to 2, and the warm-up period was configured for 3 epochs.For SAM, we froze the visual encoder and prompt encoder. In the weakly-supervised setting, the mask decoder was also frozen. In contrast,when training the mask decoder, we utilized the AdamW optimizer with a learning rate of $1 \times 10^{-4}$, optimizing the BCELoss objective function. We trained the model for 20 epochs, with a batch size of 1, based on varying amounts of data. The setup was run on two RTX 3090 GPU.
\subsection{Evaluation Settings}

\textbf{Dataset.}Our experiments are conducted on the benchmark dataset for referring remote sensing image segmentation (RRSIS-D)\cite{liu2024rotated}. RRSIS-D contains 17,402 images, each accompanied by corresponding masks and referring expressions. The dataset includes semantic labels across 20 categories, supplemented by 7 attributes. RRSIS-D also offers greater flexibility in mask resolution and exhibits significant scale variability. The dataset presents a challenging task due to the large-scale variations and numerous small targets, requiring prediction across diverse scales. To enhance compatibility with natural RIS models, the dataset uses the RefCOCO annotation format to improve its usability.

\textbf{Metrics.}We use Overall Intersection over Union (oIoU), Mean Intersection over Union (mIoU), and Precision@X (P@X) as evaluation metrics. oIoU is calculated by dividing the total intersection pixels of all test images by the total union pixels. Precision@X computes the percentage of test samples where the predicted IoU scores exceed the threshold X. mIoU is obtained by calculating the Intersection over Union (IoU) for each category and then averaging the IoUs across all categories.

\begin{table*}[t]  
    \centering
    \caption{Comparison with state-of-the-art methods on RRSIS-D dataset. 
     R-101 and Swin-B represent ResNet-101 and base Swin Transformer models, respectively. $^{*}$ denotes weakly supervised training setting (using only bounding box data).
     The best result is \textbf{bold}.}
    \renewcommand{\arraystretch}{1} 
    \begin{tabular}{cccccccccc}
    \toprule
        \large Method & \large $E_v$ & \large $E_t$ & \large $Pr@0.5$ & \large $Pr@0.6$ & \large $Pr@0.7$ & \large $Pr@0.8$ & \large $Pr@0.9$ & \large oIoU & \large MIoU \\ 
    \midrule
        RRN\cite{li2018referring} & R-101 & LSTM & 51.08 & 42.14 & 32.74 & 21.67 & 6.47 & 66.47 & 45.62 \\ 
        CMSA\cite{ye2019cross} & R-101 & None & 55.34 & 46.45 & 37.53 & 25.46 & 8.17 & 69.49 & 48.52 \\ 
        LSCM\cite{hui2020linguistic} & R-101 & LSTM & 56.12 & 46.28 & 37.71 & 25.23 & 8.29 & 69.08 & 49.97 \\ 
        CMPC\cite{huang2020referring} & R-101 & LSTM & 55.84 & 47.43 & 36.93 & 25.49 & 9.23 & 69.27 & 49.28 \\ 
        BRINet\cite{hu2020bi} & R-101 & LSTM & 56.96 & 48.79 & 39.15 & 27.11 & 8.74 & 69.78 & 49.68 \\ 
        CMPC+\cite{liu2021cross} & R-101 & LSTM & 57.68 & 47.54 & 36.98 & 24.31 & 7.81 & 68.63 & 50.28 \\ 
    \midrule
        LGCE\cite{yuan2024rrsis} & Swin-B & BERT & 67.69 & 61.55 & 51.47 & 39.64 & 23.35 & 76.37 & 59.41 \\ 
        LAVT\cite{yang2022lavt} & Swin-B & BERT & 69.57 & 63.68 & 53.31 & 41.65 & 24.98 & 77.22 & 61.43 \\ 
        CARIS\cite{liu2023caris} & Swin-B & BERT & 71.61 & 64.67 & 54.17 & 42.78 & 23.89 & 77.41 & 63.35 \\
        RMSIN\cite{liu2024rotated} & Swin-B & BERT & 74.28 & 67.25 & 55.94 & 42.57 & 24.58 & 77.53 & 64.22 \\ 
        
    \midrule
        PSLG-SAM$^{*}$ & \multicolumn{2}{c}{SAM+LQVG} & 78.92  & 70.44 & 58.80 & 45.66 & 27.92 & 75.40 & 67.41 \\ 
        PSLG-SAM & \multicolumn{2}{c}{SAM+LQVG} & \textbf{83.48} & \textbf{77.85} & \textbf{66.25} & \textbf{49.84} & \textbf{30.22} & \textbf{77.90} & \textbf{70.61} \\ 
    \bottomrule
    \end{tabular}
    
    \label{tab:com_RMSIN}
\end{table*}
\begin{table*}  
     \caption{Comparison with state-of-the-art methods on RRSIS-M dataset.R-101 and Swin-B represent ResNet-101 and base Swin Transformer models, respectively. $^{*}$ denotes weakly supervised training setting (using only bounding box data).
     The best result is \textbf{bold}.}
    \begin{tabular}{cccccccccc}
    \toprule
        \large Method & \large $E_v$ & \large $E_t$ & \large $Pr@0.5$ & \large $Pr@0.6$ & \large $Pr@0.7$ & \large $Pr@0.8$ & \large $Pr@0.9$ & \large oIoU & \large MIoU \\ 
    \midrule
        RRN\cite{li2018referring} & R-101 & LSTM & 50.22 & 41.52 & 30.04 & 24.13 & 17.76 & 53.27 & 42.73 \\ 
        CMSA\cite{ye2019cross} & R-101 & None & 53.48 & 45.32 & 34.83 & 25.48 & 19.11 & 56.39 & 45.41 \\ 
        LSCM\cite{hui2020linguistic} & R-101 & LSTM & 54.14 & 45.92 & 35.89 & 25.17 & 19.57 & 55.08 & 45.11 \\ 
        CMPC\cite{huang2020referring} & R-101 & LSTM & 55.02 & 45.19 & 36.08 & 26.87 & 18.73 & 55.57 & 46.71 \\ 
        BRINet\cite{hu2020bi} & R-101 & LSTM & 55.68 & 46.07 & 36.58 & 26.10 & 19.78 & 55.94 & 46.22 \\ 
        CMPC+\cite{liu2021cross} & R-101 & LSTM & 55.72 & 46.26 & 35.71 & 27.69 & 18.31 & 55.27 & 46.83 \\ 
    \midrule
        LGCE\cite{yuan2024rrsis} & Swin-B & BERT & 68.19 & 63.94 & 55.26 & 50.66 & 38.78 & 68.74 & 62.95 \\ 
        CARIS\cite{liu2023caris} & Swin-B & BERT & 70.10 & 64.58 & 60.07 & 53.26 & 39.71 & 69.84 & 63.87 \\ 
        LAVT\cite{yang2022lavt} & Swin-B & BERT & 68.57 & 64.26 & 61.14 & 53.08 & 39.67 & 69.48 & 63.75 \\ 
        RMSIN\cite{liu2024rotated} & Swin-B & BERT & 69.02 & 64.76 & 59.78 & 53.42 & 39.76 & 69.86 & 63.98 \\ 
        
    \midrule
        PSLG-SAM$^{*}$ & \multicolumn{2}{c}{SAM+LQVG} & 71.09  & 65.31 & 62.37 & 52.28 & 38.74 & 68.26 & 63.31 \\ 
        PSLG-SAM & \multicolumn{2}{c}{SAM+LQVG} & \textbf{74.82} & \textbf{71.38} & \textbf{66.58} & \textbf{59.33} & \textbf{43.30} & \textbf{71.16} & \textbf{67.04} \\ 
    \bottomrule
    \end{tabular}
     \label{tab:com_rrsisd}
\end{table*}

\subsection{Comparison with state-of-the-art RIS methods}
In our experiment, we compared the performance of PSLG-SAM  with existing state-of-the-art reference image segmentation methods, as shown in \autoref{tab:com_RMSIN} and \autoref{tab:com_rrsisd}. Our method was tested under two settings: weakly supervised experiments trained only with bounding box data and regular supervised training setting with mask-lvel label. From the table, we observed that the method using ResNet-101 as the image encoder and LSTM as the text encoder performed poorly on the RRSIS dataset. In contrast, methods based on Swin-Transformer and BERT generally improved by about 10\%. This improvement can be attributed to the pre-trained vision and language encoders, which brought additional knowledge and alleviated the issue of insufficient training data.

Among all the compared methods, our approach, which introduces segment anything model, significantly outperformed existing methods in terms of precise and mIoU. In the weakly supervised setting on the RRSIS-D dataset, compared to the state-of-the-art RMSIN method, our approach achieved a 3.2\% improvement in mIoU and a 2.1\% difference in oIoU. This indicates that our framework is better at capturing small target objects and accurately segmenting them. The lower oIoU, however, suggests that without fine-tuning on mask data, the decoder struggles with distinguishing the boundaries between targets and backgrounds in unseen remote sensing images.

After training with mask data, our method achieved SOTA performance on every metric, with a particularly significant 6.4\% improvement in mIoU and 0.5\% improvement in oIoU on the RRSIS-D dataset. On the RRSIS-M dataset, our method achieves a 3\% higher mIoU (mean Intersection over Union) and 1.3\% higher oIoU (object Intersection over Union) compared to existing approaches.This demonstrates the effectiveness of the proposed framework.

\subsection{Ablation study}
We conducted ablation experiments on the test subset of RRSIS-D to evaluate the effectiveness of our proposed method. The experiments primarily include two parts: (1) Ablation of key components, and (2) Comparative experiments with varying amounts of training data.
\subsubsection{Ablation for key components}
To validate the effectiveness of the CFPG module and the ABC strategy, we conducted ablation experiments involving all possible combinations of these two components. As shown in \autoref{ablation}, introducing the CFPG module significantly improves both prcise and mIoU, while using the MBO strategy alone results in a substantial increase in oIoU. The combination of both modules demonstrates synergistic enhancement, achieving the best performance across all evaluation metrics, with notable improvements of 5.6\% in oIoU and 2.1\% in mIoU. These results indicate that the CFPG module and MBO strategy enable the decoder to produce more accurate segmentation masks and refined mask outputs, further confirming their effectiveness in enhancing overall segmentation performance.
\begin{table}[h]
    \caption{Ablation study to evaluate the contributions of key components.}
    \begin{tabular}{cc|ccccc}
        \toprule
        CFPG & MBO & P@0.5 & P@0.7 & P@0.9 & oIoU & mIoU \\
        \midrule
          &  & 76.87 & 54.98 & 23.53 & 69.86 & 65.31 \\
        \checkmark &  & 80.29 & 58.23 & 26.77 & 74.08 & 67.34 \\
          & \checkmark & 77.07 & 57.81 & 26.79 & 75.24 & 66.58 \\
        \checkmark & \checkmark & 78.90 & 58.80 & 27.92 & 75.40 & 67.41 \\
        \bottomrule
    \end{tabular}
    \label{ablation}
\end{table}
\begin{table}[h]
    \caption{Comparative experiments with varying amounts of training data.}
    \begin{tabular}{c|ccccc}
    \toprule
    percent & Pr@0.5 & Pr@0.7 & Pr@0.9 & oIoU & mIoU \\ \midrule
    0\%    & 78.90    & 58.80    & 27.92    & 75.40  & 67.41  \\ 
    10\%   & 83.34   & 64.90   & 29.10    & 77.36 & 70.13 \\ 
    50\%   & 83.57   & 64.75   & 29.99   & 78.08 & 70.35 \\ 
    100\%  & 83.48   & 66.25   & 30.22   & 77.90 & 70.67 \\ 
    \bottomrule
    \end{tabular}
    
    \label{tab:ablation_experiment}
\end{table}
\subsubsection{Comparation of different training data sizes}
We fine-tuned the SAM mask decoder under different amounts of data to validate the effectiveness of our framework in scenarios where there is a lack of sufficient labeled samples. As shown in \autoref{tab:ablation_experiment}, even when using only 10\% of the data for training, the performance improvement is still quite significant. Upon examining the next two rows, it can be observed that further increasing the amount of mask-level data leads to diminishing returns in performance improvement. This is because the SAM mask decoder has relatively few parameters, and with the encoder frozen, it does not require extensive data. When transferred to scenarios with large domain discrepancies, our model achieves good results with only a small amount of mask annotations.
\begin{figure*}[!ht]
    \centering
    \includegraphics[width=0.95\linewidth]{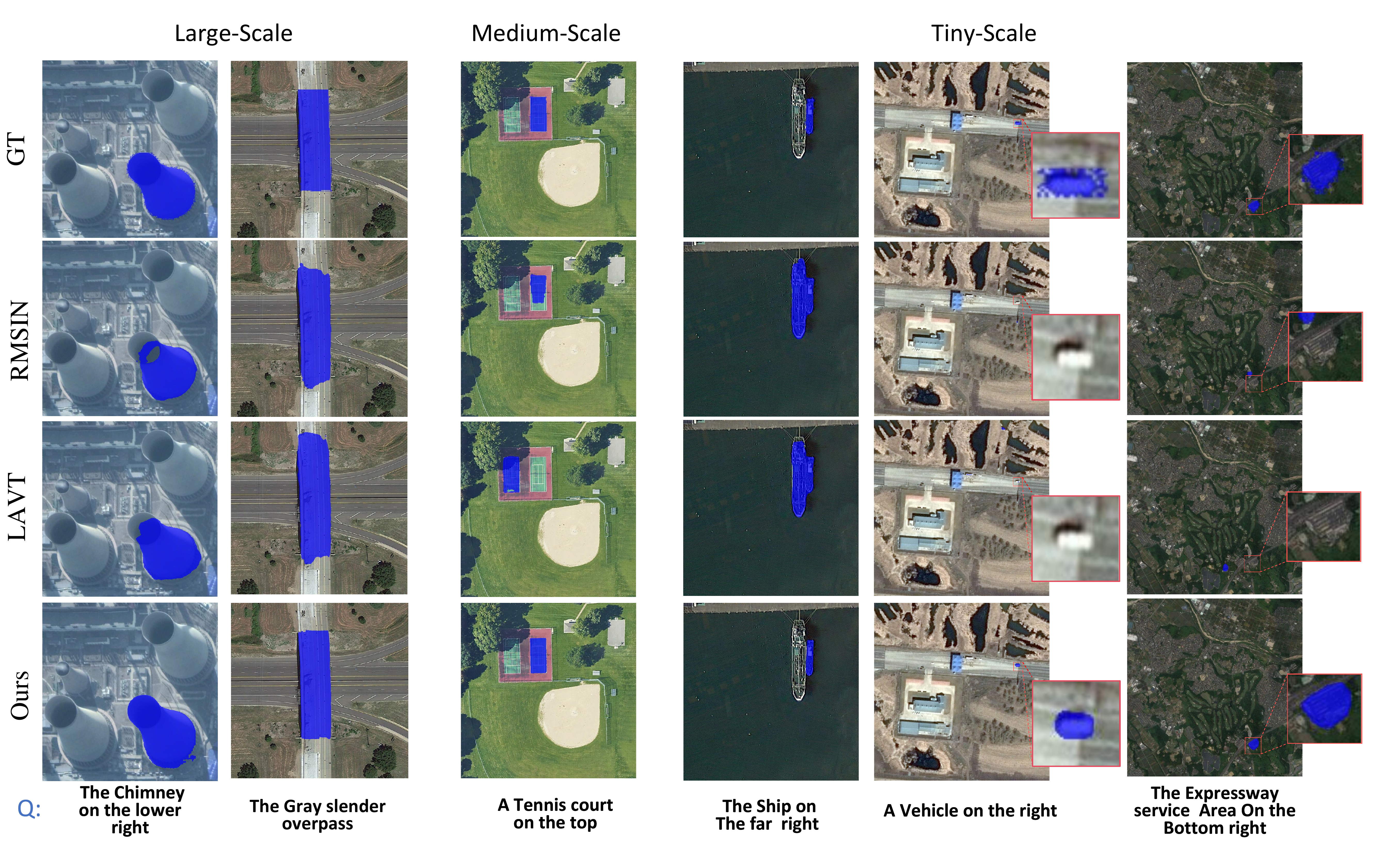}
    \caption{Qualitative comparisons between proposed method and the previous methods RMSIN and LAVT. 
    }
    \label{Visualizations}
\end{figure*}
\begin{figure*}[!ht]
    \centering
    \includegraphics[width=0.7\linewidth]{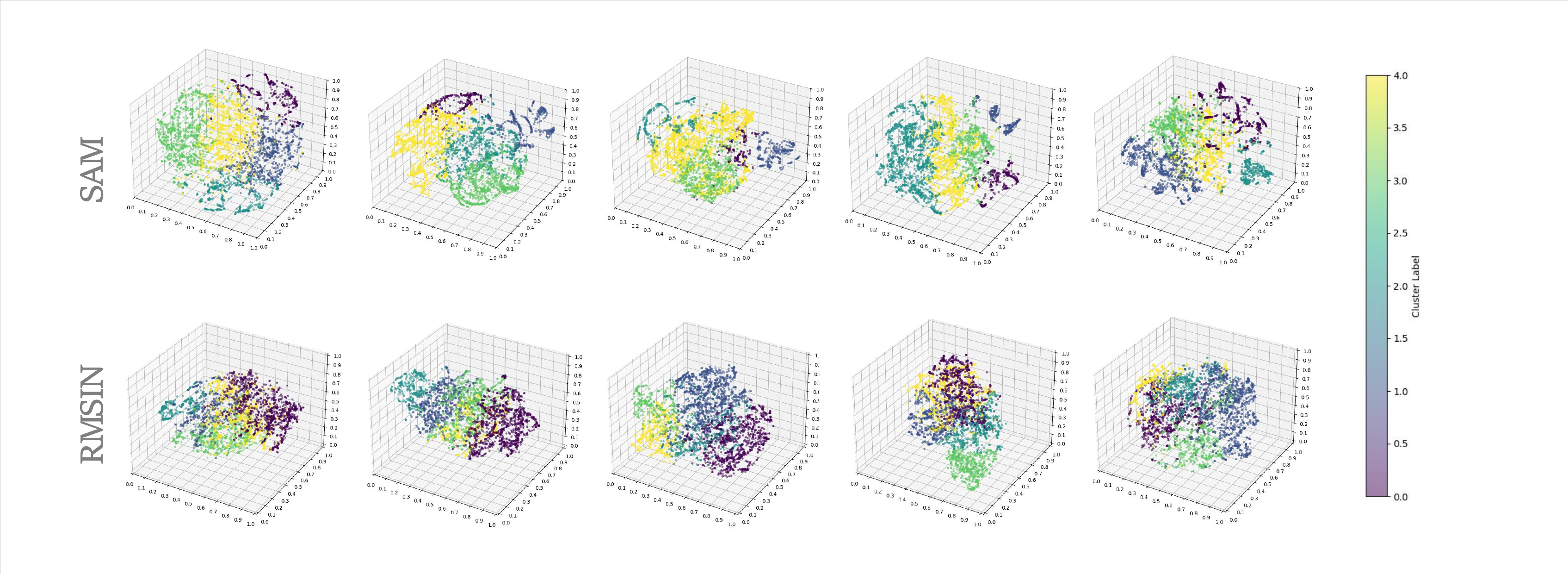}
    \caption{t-SNE visualization of Encoder output features between proposed method and the previous methods RMSIN. The output features are divided into five categories for t-SNE visualization, with the displayed images being randomly selected.
    }
    \label{Visualization}
\end{figure*}

\subsection{Visualization}
\subsubsection{Quantitative Results }
To gain deeper insights into PSLG-SAM, we visualized the qualitative comparisons between the predictions of our model and those of RMSIN and LAVT methods. As illustrated in \autoref{Visualizations}, the introduction of the visual localization model significantly improves the detection accuracy for small objects within our framework. Furthermore, when examining the segmentation results for larger objects, our framework effectively reduces segmentation errors compared to other methods.
\subsubsection{Visualization of Encoder Output }
To assess the fine-grained discriminative ability of the image encoder in complex scenes, we performed a visualization analysis of the encoder's output. We fixed the number of object categories in the image and extracted the feature outputs from the last layer of the SAM encoder (for RMSIN, we selected the second-to-last layer features after multi-scale interaction to ensure approximate dimensionality). These high-dimensional features were then reduced using the t-SNE method. We conducted experiments with the number of categories ranging from 5 to 10. The experimental results indicate that, in most scenes, the SAM encoder effectively captures subtle differences in the image and can distinguish between different object types. Specifically, compared to the RMSIN method,as shown in \autoref{Visualization}, the feature outputs of SAM show higher density within the same category, while the distances between different categories are larger. This analysis confirms the superior fine-grained discriminative ability of the SAM encoder.
\section{Conclusion}
In this paper, we introduce the PSLG-SAM framework, which effectively addresses the challenges of Reference Remote Sensing Image Segmentation (RRSIS) by decomposing the task into two distinct stages: coarse localization and fine-grained segmentation. Our approach leverages visual grounding to approximate object locations and utilizes SAM’s boundary segmentation capabilities for precise object delineation. The framework also incorporates an unsupervised centroid generator for improved prompt accuracy and a mask boundary optimization strategy to refine segmentation results. 

Experimental validation on the RRSIS-D dataset demonstrates significant performance improvements, showcasing the framework’s effectiveness in handling complex remote sensing images with minimal training data requirements. PSLG-SAM represents a scalable, efficient solution for RRSIS tasks, offering promising potential for broader applications in remote sensing.
\bibliographystyle{ACM-Reference-Format}
\bibliography{sample-base}
\end{document}